\documentclass{article}

\PassOptionsToPackage{numbers,sort&compress}{natbib}
\usepackage[preprint]{neurips_2026}

\usepackage[utf8]{inputenc} 
\usepackage[T1]{fontenc}    
\usepackage{hyperref}       
\usepackage{url}            
\usepackage{booktabs}       
\usepackage{tabularx}       
\usepackage{array}          
\usepackage{amsfonts}       
\usepackage{nicefrac}       
\usepackage{microtype}      
\usepackage{graphicx}       
\usepackage{float}          
\usepackage{caption}        
\usepackage{xcolor}         
\usepackage[normalem]{ulem} 
\usepackage{tcolorbox}      
\tcbuselibrary{breakable,skins,listings}
\usepackage{amsmath}        
\usepackage{fvextra}        
\usepackage{clrscode3e}     
\usepackage{fontawesome5}   

\definecolor{gain}{HTML}{22863a}
\definecolor{loss}{HTML}{cb2431}
\newcommand{\gdelta}[1]{\textcolor{gain}{(#1)}}
\newcommand{\rdelta}[1]{\textcolor{loss}{(#1)}}
\newcommand{\evaltabheader}{%
  \toprule
  & base & SFT & RL & SFT$\rightarrow$RL & O4-mini & GPT-5 & Opus 4.6 \\
  \midrule
}

\definecolor{warnred}{HTML}{cb2431}

\newtcolorbox{contribbox}{
  colback=black!2,
  colframe=black!20,
  boxrule=0.4pt,
  arc=2pt,
  left=7pt,
  right=7pt,
  top=4pt,
  bottom=4pt,
  before skip=5pt,
  after skip=5pt,
}

\newtcolorbox{rqbox}{
  colback=blue!4,
  colframe=blue!35,
  boxrule=0.4pt,
  leftrule=2.5pt,
  arc=0pt,
  left=8pt,
  right=8pt,
  top=5pt,
  bottom=5pt,
  before skip=8pt,
  after skip=8pt,
}

\newtcolorbox{promptblock}{
  enhanced,
  colback=black!2,
  colframe=black!18,
  boxrule=0.4pt,
  arc=2pt,
  left=7pt,
  right=7pt,
  top=5pt,
  bottom=5pt,
  breakable,
  before skip=10pt,
  after skip=6pt,
}

\DefineVerbatimEnvironment{promptverbatim}{Verbatim}{%
  breaklines=true,
  breakanywhere=true,
  breaksymbolleft={},
  breaksymbolright={},
  fontsize=\normalsize,
}

\definecolor{anatSystem}{HTML}{1565C0}
\definecolor{anatUser}{HTML}{7B1FA2}
\definecolor{anatRubric}{HTML}{00838F}
\definecolor{anatText}{HTML}{C2185B}
\definecolor{anatAnswer}{HTML}{2E7D32}
\newcommand{\anatomysubsection}[3]{%
  \subsection{\textcolor{#1}{#2}}%
  \label{#3}%
}
\newcommand{\promptvariant}[1]{%
  \paragraph{#1}%
  \mbox{}\par\vspace{6pt}%
}

\newcommand{\promptwhere}{%
  \par\vspace{4pt}%
  \noindent where:%
}
\newcommand{\mutationlabel}[1]{%
  \noindent\textbf{#1:}%
  \par\vspace{2pt}%
}

\definecolor{diffdel}{HTML}{CF222E}
\definecolor{diffadd}{HTML}{1A7F37}
\newcommand{\diffdel}[1]{\textcolor{diffdel}{\sout{#1}}\ }
\newcommand{\diffadd}[1]{\textcolor{diffadd}{#1}\ }

\definecolor{pipeX}{HTML}{7B1FA2}
\definecolor{pipeY}{HTML}{6D4C41}
\definecolor{pipeO}{HTML}{E65100}
\definecolor{pipeM}{HTML}{1565C0}
\newcommand{\varx}{\ensuremath{x}}
\newcommand{\vary}{\ensuremath{y}}
\newcommand{\varo}{\ensuremath{o}}

\newcommand{\varm}{\ensuremath{m}}
\newcommand{\vars}[1]{\ensuremath{s_{#1}}}

\usepackage{listings}
\definecolor{jsonstr}{HTML}{032F62}
\definecolor{pykw}{HTML}{0033B3}

\lstdefinelanguage{MutationPython}{%
  language=Python,
  basicstyle=\ttfamily\footnotesize,
  showstringspaces=false,
  breaklines=true,
  breakatwhitespace=false,
  columns=flexible,
  keepspaces=true,
  frame=none,
  aboveskip=0pt,
  belowskip=0pt,
  keywordstyle=\color{pykw},
  stringstyle=\color{jsonstr},
}

\newenvironment{mutationplain}{%
  \begin{promptblock}%
  \begingroup\ttfamily\footnotesize\frenchspacing\spaceskip=0.33em plus 0.12em minus 0.08em\relax
}{%
  \endgroup\end{promptblock}%
}

\newcommand{\picoq}[1]{"#1"}
\newcommand{\picoline}[2]{\textbf{#1}: #2\par\vspace{3pt}}
\newcommand{\argline}[2]{\textbf{#1}: #2\par\vspace{2pt}}

\newcommand{\mutationexample}[2]{%
  \noindent Split=#2, ID=#1.\par\vspace{4pt}%
}

\newcommand{\mutationsubsection}[3]{%
  \subsection{\texorpdfstring{\href{#2}{#1}}{#1}}%
  \label{#3}%
}

\newtcblisting{mutationpython}{%
  enhanced,
  colback=black!2,
  colframe=black!18,
  boxrule=0.4pt,
  arc=2pt,
  left=7pt,
  right=7pt,
  top=5pt,
  bottom=5pt,
  breakable,
  before skip=10pt,
  after skip=6pt,
  listing only,
  listing options={%
    language=MutationPython,
    basicstyle=\ttfamily\footnotesize,
    showstringspaces=false,
    breaklines=true,
    breakatwhitespace=false,
    columns=flexible,
    keepspaces=true,
    frame=none,
    aboveskip=0pt,
    belowskip=0pt,
    keywordstyle=\color{pykw},
    stringstyle=\color{jsonstr},
  },
}

\newtcolorbox{algorithmbox}[1][]{%
  enhanced,
  colback=blue!3,
  colframe=blue!35,
  boxrule=0.4pt,
  leftrule=2.5pt,
  arc=0pt,
  left=8pt,
  right=8pt,
  top=6pt,
  bottom=6pt,
  breakable,
  before skip=8pt,
  after skip=8pt,
  fonttitle=\bfseries,
  title={#1},
}

\newcommand{\rqone}{How often are judge disagreements operational in biomedical tasks?}
\newcommand{\rqtwo}{What training regime, if any, works best for small LLM judges in biomedical tasks?}
\newcommand{\rqthree}{Can deterministic data generation model difficulty in partially correct biomedical tasks?}

\newcommand{\printrqonebox}{%
  \begin{rqbox}
  \textbf{RQ1:} \textit{\rqone}
  \end{rqbox}%
}

\newcommand{\printrqtwobox}{%
  \begin{rqbox}
  \textbf{RQ2:} \textit{\rqtwo}
  \end{rqbox}%
}

\newcommand{\printrqthreebox}{%
  \begin{rqbox}
  \textbf{RQ3:} \textit{\rqthree}
  \end{rqbox}%
}

\title{Beyond Correctness: Validity-Oriented Evaluation of Biomedical LLM Judges}
\workshoptitle{First Workshop on Reliable Evaluation for Language Models (JUDGe)}

\author{%
  Rodrigo de Oliveira \\
  Independent Researcher\thanks{This work was largely carried out while the author was employed at IQVIA.}, London, UK \\
  \texttt{rdeoliveira.phd@gmail.com} \\
  \And
  Federico Pittino \\
  IQVIA, Barcelona, Spain\\
  \texttt{federico.pittino@iqvia.com} \\
  \And
  James Gwinnutt \\
  IQVIA, Reading, UK\\
  \texttt{james.gwinnutt@iqvia.com} \\
  \And
  Jay Nanavati \\
  IQVIA, Cambridge, UK\\
  \texttt{jay.nanavati@iqvia.com} \\
}

\begin{document}

\maketitle
\begin{abstract}
We propose a scalable, validity-oriented pipeline for evaluating biomedical LLM judges when high-quality human judgments are scarce. First, we augment existing human-labelled biomedical benchmarks with deterministic, metric-grounded mutations that produce auditable preference pairs. Second, we evaluate judges beyond aggregate correctness using three deployment-relevant dimensions: correctness against metric-derived gold labels, robustness under repeated stochastic sampling, and compliance with the requested output format. We use this pipeline to assess Llama-3.1-8B-Instruct under four regimes: (1)~\emph{base}, using the instruct model as is; (2)~\emph{SFT}, distillation-based supervised fine-tuning only; (3)~\emph{RL}, GRPO-based reinforcement learning only; and (4)~\emph{SFT$\rightarrow$RL}, SFT followed by RL. The base and single-stage regimes struggle on structured medical discrimination such as PICO extraction and clinical calculations, whereas SFT$\rightarrow$RL performs best across correctness, compliance, and robustness; gains concentrate on decomposable tasks (PICO, MedCalc), at times matching or outperforming frontier models.
\end{abstract}

\section{Introduction}

LLM-as-a-judge is increasingly used to scale evaluation beyond human ratings~\cite{mtbench}. In high-stakes settings such as healthcare, however, judges must remain valid in real-world use. Biomedical applications---from structured extraction and clinical summarization~\cite{npjclinjudges} to clinical interviews~\cite{clinichat} and exam-style question answering~\cite{medqa,bioasq2025}---require fine-grained discrimination of PICO spans, calculation steps, and subtle distractors.

Most existing work treats agreement with human or expert judgments as the primary quality indicator~\cite{mtbench,rewardbench,judgebench}, giving less attention to operational failures. A disagreement may reflect a substantive error, but may also arise from verdict instability or a non-parseable response~\cite{offsetbias,sage}. We therefore introduce a three-axis protocol measuring criterion-linked correctness, robustness, and compliance, and ask:
\printrqonebox

We compare small and frontier judges across biomedical tasks. Frontier models provide a strong ceiling, but fine-tuning small models may offer a more cost-effective route to specialized evaluators. Recent work trains thinking judges with supervised fine-tuning or reinforcement learning~\cite{j1,j4r,thinkj,judgelrm,rmr1,glider}, typically via GRPO~\cite{deepseekmath}; we therefore ask:
\printrqtwobox

Training and evaluating these judges requires preference data--i.e. pairs of chosen/rejected responses--that would be costly to obtain through new human-rating campaigns. We therefore follow the approach of deterministic synthetic-data generation with verifiable ground truth~\cite{flexece,npjclinjudges,clinichat,medqa,bioasq2025}. In real-world pipelines, strong generators often produce very good responses, thereby increasing the difficulty for LLM judges: it is arguably easier to tell a good from a bad response, than a good from a great response. As such, it is highly desirable to adjust the difficulty of synthetic pairs at the point of generation, motivating our final question:
\printrqthreebox

Thus, our contributions are three-fold:
\begin{enumerate}
  \item A three-axis, validity-oriented \textbf{evaluation protocol} for biomedical LLM judges, combining criterion-linked correctness with robustness under resampling and compliance.
  \item A deterministic and scalable method for augmenting biomedical benchmarks with auditable, metric-grounded \textbf{synthetic preference data} for judge training and evaluation.
  \item Finally, an \textbf{empirical study of} four distinct \textbf{training} regimes of \textbf{small LLMs} using the proposed protocol and augmented data.
\end{enumerate}

\section{Method}

\subsection{Biomedical tasks}
\label{sec:biomedical-tasks}

We train and evaluate on five biomedical tasks\footnote{PICO --- \url{https://huggingface.co/datasets/bigbio/ebm_pico}; MedCalc --- \url{https://huggingface.co/datasets/ncbi/MedCalc-Bench-v1.2}; MedBullets --- \url{https://huggingface.co/datasets/super-dainiu/medagents-benchmark}; MedQA --- \url{https://huggingface.co/datasets/super-dainiu/medagents-benchmark}; PubMedQA --- \url{https://huggingface.co/datasets/super-dainiu/medagents-benchmark}.} grouped into two families by response structure.
\emph{Decomposable} tasks---PICO span extraction and MedCalc---require judging partially correct outputs: whether extracted trial elements or clinical calculations are right.
\emph{Atomic} tasks---MedBullets, MedQA, and PubMedQA---require judging all-or-nothing exam-style answers where one option is fully correct.

\subsection{Synthetic data generation}
\label{sec:synthetic-data}

Every preference label is traceable to ground truth and a task metric.
Algorithms~\hyperref[alg:synth-candidates]{1} and~\hyperref[alg:synth-pair]{2} in Appendix~\ref{app:mutations} implement candidate generation, pair selection, and the suite-specific operator~\varo\ and metric~\varm\ for each benchmark.
For each source example with clinical text~\varx\ and ground truth~\vary, a task-specific mutation operator~\varo\ generates $K$ candidate responses:
\begin{equation}
  \label{eq:generation}
  \{c_1,\ldots,c_K\} = o(x,y).
\end{equation}
On PICO, \varo\ adds or deletes spans; on MedCalc, it perturbs numeric inputs; on atomic tasks, it substitutes multiple-choice options (Appendix~\ref{app:mutations}).

Each candidate is scored against ground truth and ranked by quality (Algorithm~\hyperref[alg:synth-candidates]{1}):
\begin{equation}
  \label{eq:scoring}
  s_k = m(c_k, y),
\end{equation}
Here, \varm\ is macro-F1 on PICO, MSE on MedCalc, and binary correctness on atomic tasks.

From the ranked pool, preprocessing selects one valid chosen--rejected pair per source example (Algorithm~\hyperref[alg:synth-pair]{2}).
Each retained \emph{pair} is presented in slots $A$ and $B$ with gold verdict $l \in \{A,B\}$ naming the preferred slot:
\begin{equation}
  \label{eq:pair}
  l =
  \begin{cases}
    A & \text{if } s_A > s_B, \\
    B & \text{otherwise},
  \end{cases}
\end{equation}
where $s_A$ and $s_B$ are the candidate scores.
On decomposable tasks, \emph{pairs} are drawn from the ranking with integer rank gap
\begin{equation}
  \label{eq:rank-gap}
  \delta = j - i,
\end{equation}
where $i < j$ are the rank positions (smaller $\delta$ = harder \emph{pair}); we stratify decomposable pairs into three per-example $\delta$ tertiles and drop pairs whose scores are too close.
On atomic tasks, \emph{pairs} contrast correct and incorrect options without rank-gap stratification.

\paragraph{SFT data.}
A frontier teacher (Claude Opus~4.6\footnote{\url{https://www-cdn.anthropic.com/14e4fb01875d2a69f646fa5e574dea2b1c0ff7b5.pdf}\label{fn:opus-system-card}}) samples judge generation $g$, which receives a deterministic quality score $Q$:
\begin{equation}
  \label{eq:sft-quality}
  Q(g,l) = f_{\mathrm{fmt}}(g) + f_{\mathrm{parse}}(g) + f_{\mathrm{align}}(g,l) \in \{0,1,2,3\},
\end{equation}
where $f_{\mathrm{fmt}}$ and $f_{\mathrm{parse}}$ require valid J1 judge format and a parseable verdict or score.
The SFT corpus distills both pairwise verdict and pointwise score prompts from the same pairs.
For pairwise examples, $f_{\mathrm{align}}{=}1$ iff the teacher's A/B verdict matches the metric-derived gold verdict $l$.
For pointwise examples, the teacher scores each candidate in a preference pair independently; $f_{\mathrm{align}}{=}1$ iff its score on the chosen candidate exceeds its score on the rejected candidate.

\paragraph{RL data.}
RL uses the same pairs and gold verdicts to provide verifiable J1-style rewards~\cite{j1}; no teacher completions are stored.
Because judges are sensitive to candidate order~\cite{mtbench}, we randomly swap slots during training and evaluation.

The pipeline scales with existing benchmarks rather than new rating campaigns, and its fixed operators and metrics make every pair auditable and replayable.

\subsection{Training}
\label{sec:training}

All experiments use Llama-3.1-8B-Instruct on synthetic pairs from the five-suite pipeline: ${\sim}$27K training preference pairs for RL and SFT$\rightarrow$RL, and ${\sim}$48K distilled teacher completions with $Q{=}3$ (pairwise and pointwise) for SFT; ${\sim}$3K pairs are held out for meta-evaluation (§~\ref{sec:meta-evaluation}).
Judge prompt templates are in Appendix~\ref{app:prompts}.
We implement SFT and RL with Hugging Face's TRL package\footnote{\url{https://huggingface.co/docs/trl}}.
We compare four regimes:
\begin{itemize}
  \item \textbf{Base.}
  The Llama-3.1-8B-Instruct checkpoint is used as is, without additional training.
  \item \textbf{SFT.}
  Supervised fine-tuning on distilled teacher prompt--completion pairs that pass the maximum quality filter ($Q{=}3$), mixing pairwise and pointwise judge modes.
  The model is trained with completion-only cross-entropy on teacher completions; prompt tokens are masked from the loss.
  \item \textbf{RL.}
  Pairwise judge reinforcement learning from the instruct checkpoint, without teacher completions, following J1~\cite{j1} with verifiable pairwise rewards optimized with GRPO~\cite{deepseekmath}.
  On each preference pair the student samples a judge completion; the reward is $1$ when the parsed A/B verdict matches the gold verdict $l$, and $0$ otherwise.
  \item \textbf{SFT$\rightarrow$RL.}
  The SFT checkpoint is further tuned with the same pairwise RL stage on the same preference pairs and reward.
\end{itemize}
All tuning stages use LoRA adapters ($r{=}16$, $\alpha{=}32$) on attention and MLP projections.
SFT runs for 3 epochs with learning rate $2\times10^{-5}$, per-device batch size 2, and 8-step gradient accumulation on one NVIDIA A100 GPU.
RL runs for 1 epoch with learning rate $1\times10^{-6}$, five rollouts per prompt, and KL coefficient $\beta{=}0.01$ (J1 defaults for 8B models) on eight NVIDIA H100 GPUs.

\subsection{Meta-evaluation}
\label{sec:meta-evaluation}

We meta-evaluate every checkpoint in pairwise mode on a held-out split (${\sim}$3K comparisons across five tasks), judging each comparison three times with stochastic decoding ($\texttt{num\_repetitions}{=}3$). Correctness is agreement with the metric-derived gold verdict $l$ from Eqs.~\eqref{eq:pair}--\eqref{eq:rank-gap}; we additionally report robustness under resampling and compliance when the judge errs.
Related work studies bias, consistency, or multi-criterion assessment separately~\cite{offsetbias,sage,yescieval,npjclinjudges}; we combine correctness with two deployment-oriented diagnostics under one verifiable protocol.

Untuned frontier APIs---O4-mini\footnote{\url{https://developers.openai.com/api/docs/models/o4-mini}}, GPT-5 Chat\footnote{\url{https://developers.openai.com/api/docs/models/gpt-5-chat-latest}}, and Opus~4.6\footnotemark[\getrefnumber{fn:opus-system-card}]---serve as strong baselines.

\paragraph{Correctness.}
Pairwise correctness is majority vote over the three parsed A/B verdicts. We slice it by response type---decomposable (PICO, MedCalc) versus atomic QA---and, for decomposable tasks, by three rank-gap tertiles.
This is the axis used in most prior judge benchmarks~\cite{mtbench,rewardbench,judgebench}.

\paragraph{Robustness.}
A robust judge should produce the same rating under repeated stochastic decoding. For each comparison we derive majority-vote, all-sample, and any-sample correctness.
Let $a_{\mathrm{maj}}$, $a_{\mathrm{all}}$, and $a_{\mathrm{any}}$ denote their dataset means.
We report verdict variance as the mean whisker span
\begin{equation}
  \label{eq:robustness}
  v = \bigl|a_{\mathrm{any}} - a_{\mathrm{all}}\bigr|,
\end{equation}
averaged across datasets (equivalently, the gap between lenient any-vote and strict all-vote correctness); lower $v$ indicates less flip-flopping under resampling.
This measures resampling robustness without human re-annotation, analogous in aim to Sage~\cite{sage} but defined on repeated verdicts over fixed gold preferences.

\paragraph{Compliance.}
We separate substantive discrimination errors from format failures.
For each comparison we partition outcomes into parseable wrong verdicts (\emph{legitimate errors}) and completions without a parseable A/B verdict (\emph{invalid responses}).
This matters operationally because a non-compliant judge cannot be consumed by an automated pipeline.
Compliance failure is the invalid share among all errors,
\begin{equation}
  \label{eq:compliance}
  \rho = \frac{n_{\mathrm{invalid}}}{n_{\mathrm{invalid}} + n_{\mathrm{legitimate}}},
\end{equation}
with lower $\rho$ indicating that mistakes are more often wrong-but-parseable judgments than brittle non-responses.
Unlike bias-stress benchmarks that perturb inputs~\cite{offsetbias}, this axis measures compliance conditional on error---whether a wrong judge still returns an actionable verdict.

Together, the axes assess criterion-linked correctness, reproducibility, and actionable output.

\section{Results}

\begin{figure}[!t]
  \centering
  \includegraphics[width=\linewidth]{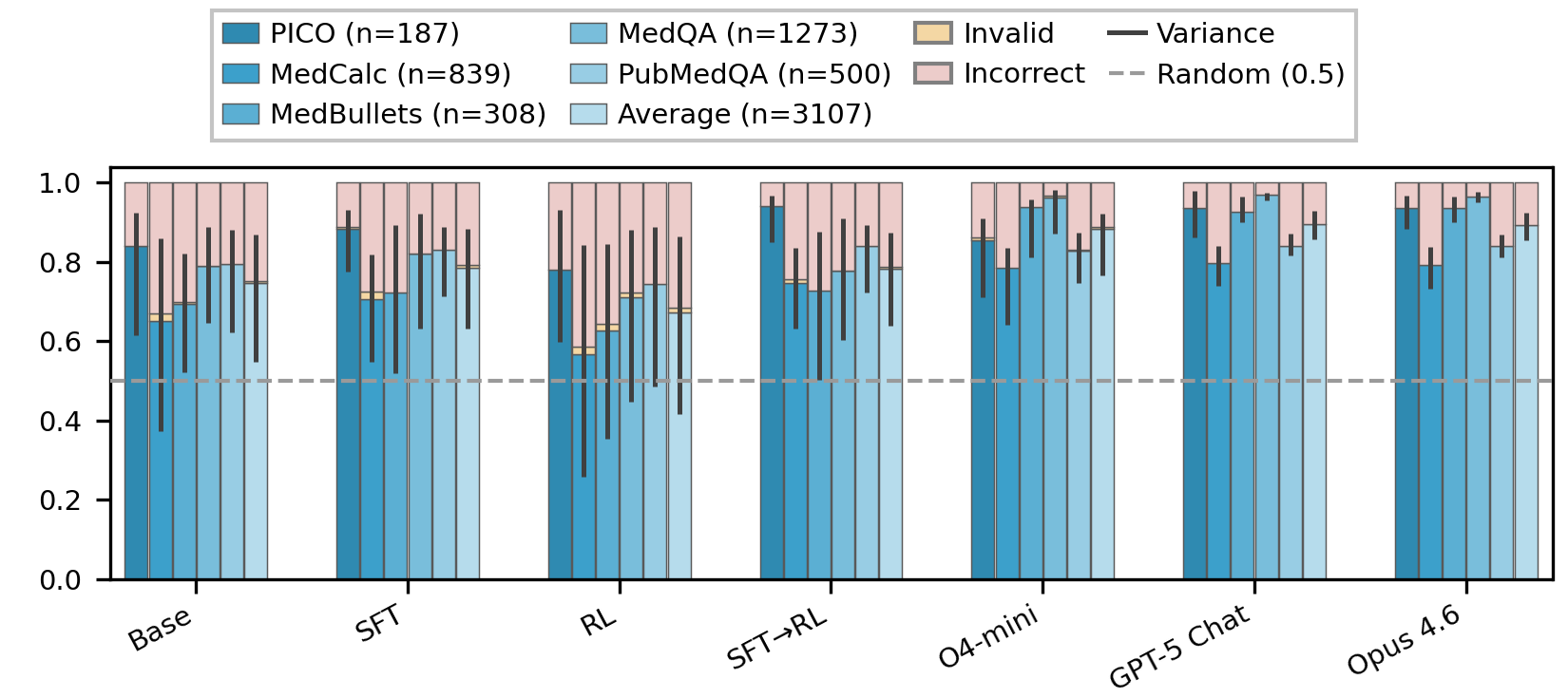}
  \caption{Response rates (correct, invalid, and incorrect) \textbf{by dataset}. Each dataset compares base, tuned, and frontier judge models ($n$ in legend). \textbf{Variance} whiskers show robustness of correct response rate (Eq.~\eqref{eq:robustness}). Exact values in Appendix~\ref{app:tables}.}
  \label{fig:judging-outcomes}
\end{figure}

\begin{figure}[!t]
  \centering
  \includegraphics[width=\linewidth]{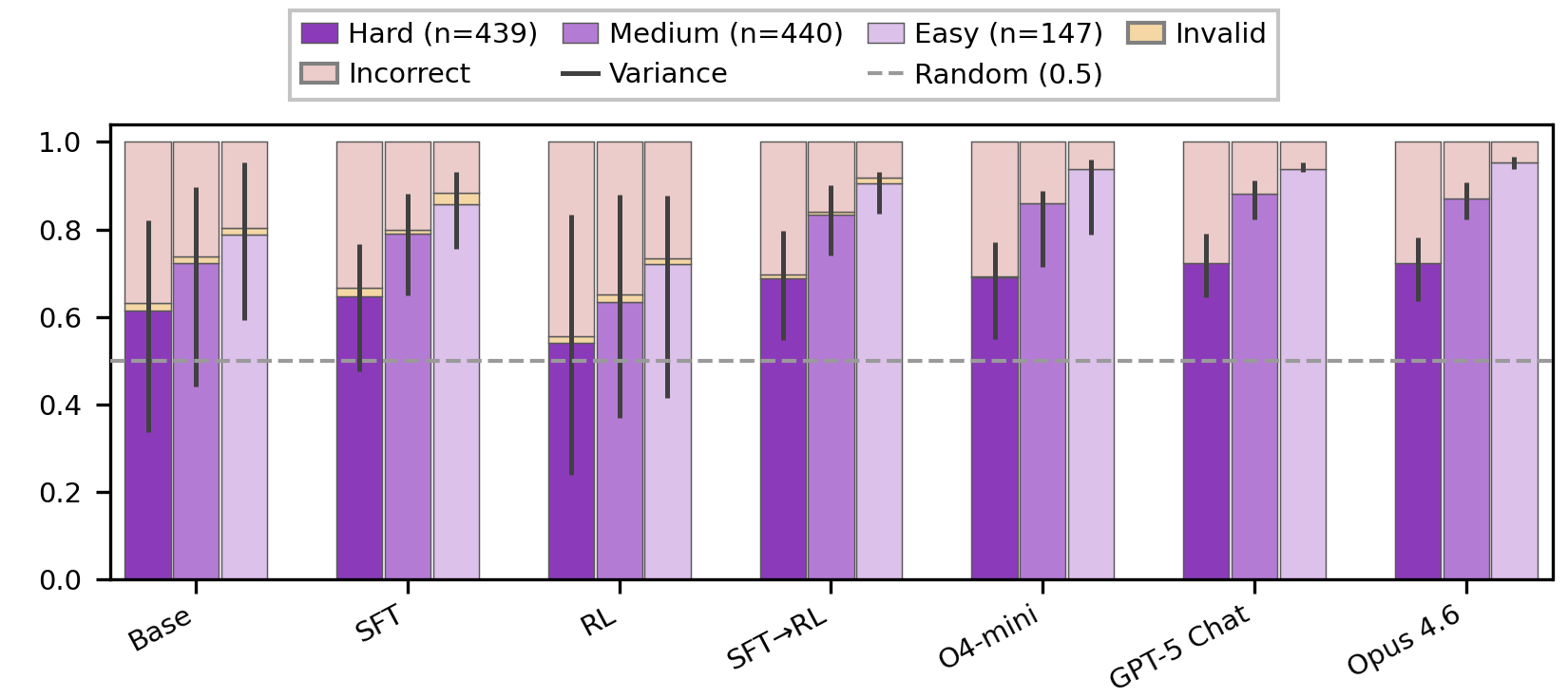}
  \caption{Response rates (correct, invalid, and incorrect) \textbf{by rank-gap difficulty}. Decomposable items only (PICO + MedCalc), stratified by \textbf{Hard}, \textbf{Medium}, and \textbf{Easy} tertiles of $\delta$ per source example (hard = smallest gaps; $n$ in legend). \textbf{Variance} whiskers show robustness of correct response rate (Eq.~\eqref{eq:robustness}). Exact values in Appendix~\ref{app:tables}.}
  \label{fig:rank-gap-difficulty}
\end{figure}

\printrqonebox

Figure~\ref{fig:judging-outcomes} shows that not every disagreement is a legitimate judging error: some failures are operational, arising from verdict instability under resampling or from non-parseable responses. Among trained regimes, SFT$\rightarrow$RL reduces the average invalid-response share to $1.6\%$ versus $3.6\%$ for RL, while its verdict variance is $0.05$ versus $0.08$ for base; the exact values are reported in Tables~\ref{tab:invalid} and~\ref{tab:variance}. Thus, operational failures are measurable and non-negligible, but can be reduced through training.

\printrqtwobox

Applying the validity-oriented evaluation protocol (§~\ref{sec:meta-evaluation}), SFT$\rightarrow$RL beats all other trained regimes on correctness, compliance, and robustness.
SFT$\rightarrow$RL achieves the highest average correctness (0.81 pairwise correctness; Figure~\ref{fig:judging-outcomes}).
Gains concentrate on decomposable items (+0.09 on PICO and MedCalc) rather than atomic QA (+0.02).
Applying RL directly to the base model without SFT is harmful ($-$0.06 versus base), which suggests biomedical judge RL needs a domain prior.
On individual suites, SFT$\rightarrow$RL matches or beats frontier judges on PICO and PubMedQA, but frontier models still lead on average and clearly dominate atomic multiple-choice discrimination.
Thus, for small biomedical judges, training order matters: RL benefits from the domain prior established by SFT.

\printrqthreebox

On decomposable tasks, rank-gap tertiles successfully model difficulty: pairwise correctness increases monotonically from hard to easy bins (Figure~\ref{fig:rank-gap-difficulty}; e.g., teach-then-incentivising: 0.69 $\rightarrow$ 0.83 $\rightarrow$ 0.90).
SFT$\rightarrow$RL gains hold at every tertile, including close-call pairs in the hardest tier.
Thus, deterministic preference construction can make synthetic evaluation more realistic by exposing judges to calibrated close-call comparisons.

\section{Limitations}

First, our experiments use one base model, Llama-3.1-8B-Instruct. We also attempted Qwen models, but many generations were non-compliant when using Hugging Face's TRL package; reporting correctness or compliance figures from those runs would therefore be misleading, so we omit them.

Second, hardware constraints limited our study to an 8B model. Larger models, such as 70B checkpoints, may behave differently and are an important direction for future work.

Third, our synthetic data pipeline is correctness-centric: it deterministically injects errors into otherwise grounded responses. The same machinery could be extended to safety-centric stress tests by deterministically injecting dangerous content into rejected responses, either instead of or alongside ordinary correctness errors.

Finally, we are aware that post-hoc calibration can improve trustworthiness in biomedical language-model applications~\cite{flexece}, but we have not yet tested calibration methods for LLM judges.

\section{Related work}

Recent work trains generative judges with RL on synthetic preferences and verifiable rewards~\cite{j1,j4r,thinkj,judgelrm,rmr1}, typically via GRPO~\cite{deepseekmath}. We do not propose a new optimizer; we transfer J1~\cite{j1} to biomedicine and test whether domain SFT should precede RL. General judge benchmarks~\cite{rewardbench,mtbench,judgebench} and tuned evaluators such as GLIDER~\cite{glider} standardize open-domain chat and reasoning; our study is domain-specific and explicitly contrasts decomposable structured tasks, where partial correctness matters~\cite{flexece}, with atomic QA. Biomedical and clinical judging~\cite{clinichat,npjclinjudges,yescieval,healthq} remains largely prompting- or SFT-based rather than pairwise judge RL~\cite{j1}.

Prior work pursues complementary judge-assessment axes: EvalBiasBench stress-tests robustness to superficial biases~\cite{offsetbias}; Sage measures local and global preference consistency without human gold~\cite{sage}; and YESciEval and clinical-judge studies employ multi-rubric or validated instruments~\cite{yescieval,npjclinjudges}.
These lines of work typically target different settings and remain separate from tuning studies.
We contribute a three-dimensional, validity-oriented evaluation protocol (correctness, robustness, compliance) on fully verifiable, metric-derived biomedical preferences.
To our knowledge, no prior work combines domain SFT, RL judge tuning, and systematic pairwise meta-evaluation along these axes across structured biomedical tasks with metric-derived reference labels.

\section{Conclusion}

In this study, we introduce a validity-oriented pipeline combining metric-grounded synthetic preferences with correctness, robustness, and compliance. We use it to compare four regimes across five biomedical benchmarks and to examine difficulty in partially correct tasks.

We answer RQ1 by showing that operational failures are distinct from legitimate judging errors: among trained regimes, SFT$\rightarrow$RL reduces the average invalid-response share to $1.6\%$ versus $3.6\%$ for RL, while its verdict variance is $0.05$ versus $0.08$ for base. These measures show why judge evaluation should extend beyond correctness alone.

We answer RQ2 by finding that SFT$\rightarrow$RL is the strongest tested regime across correctness, robustness, and compliance. Its gains concentrate on decomposable tasks, reaching $0.84$ versus $0.75$ for base, while direct RL from the base model is harmful; frontier models nevertheless remain stronger on average, especially on atomic question answering.

We answer RQ3 affirmatively: on decomposable tasks, pairwise correctness rises monotonically across rank-gap tertiles, from $0.69$ in the hardest bin to $0.90$ in the easiest for SFT$\rightarrow$RL, with its gains holding at every difficulty level. This supports using deterministic, metric-grounded pairs to study the close-call comparisons that arise when strong generators produce similarly good candidates.

Together, these results show how auditable preference data and multidimensional evaluation can make small biomedical LLM judges more reliable and operationally useful.

\bibliographystyle{unsrtnat}
\bibliography{references}

\clearpage
\appendix
\section*{Appendix}

\section{Synthetic data generation}
\label{app:mutations}

\renewcommand{\varx}{\textcolor{pipeX}{\ensuremath{x}}}
\renewcommand{\vary}{\textcolor{pipeY}{\ensuremath{y}}}
\renewcommand{\varo}{\textcolor{pipeO}{\ensuremath{o}}}
\renewcommand{\varm}{\textcolor{pipeM}{\ensuremath{m}}}

This section details the synthetic pairwise pipeline from \S\ref{sec:synthetic-data}.
For each source example with clinical text~\varx\ and ground truth~\vary, a task-specific mutation operator~\varo\ proposes $K$ candidate responses, each candidate is scored against~\vary\ with a suite-specific metric~\varm, and the pool is ranked by decreasing quality (ascending mean squared error on MedCalc).
More formally:

\hypertarget{alg:synth-candidates}{}%
\label{alg:synth-candidates}
\begin{algorithmbox}[Algorithm 1]
\begin{codebox}
\Procname{$\proc{Generate-Candidates}(x, y, o, m, K)$}
\li let $C$ and $S$ be empty arrays \Comment $C$: candidates; $S$: scores
\li \For $k \gets 1$ \To $K$ \Comment Eq.~\eqref{eq:generation}
\li     \Do
      $c \gets o(x, y)$
\li         $s \gets m(c, y)$ \Comment Eq.~\eqref{eq:scoring}
\li         append $c$ to $C$; append $s$ to $S$
  \End
\li \If{$m \isequal \text{MSE}$}
\li     \Then
      sort $C$ and $S$ so $S[1] \leq S[2] \leq \cdots \leq S[K]$
\li     \Else
      sort $C$ and $S$ so $S[1] \geq S[2] \geq \cdots \geq S[K]$
  \End
\li \Return $C$, $S$
\end{codebox}
\end{algorithmbox}

The same pipeline applies to all benchmark suites; the distinction among them reduces to the shape of~\vary, which dictates~\varo\ and~\varm\ (Table~\ref{tab:synth-by-suite}).
Decomposable tasks (PICO, MedCalc) carry structured or numeric ground truth and later use rank-gap stratification when forming pairs; atomic multiple-choice tasks use boolean correctness and contrast one correct against one incorrect candidate.
MedQA and PubMedQA follow the same atomic operator and metric as MedBullets; one QA example below suffices to illustrate that family.

\begin{table}[H]
  \centering
  \footnotesize
  \setlength{\tabcolsep}{3pt}
  \renewcommand{\arraystretch}{1.12}
  \begin{tabularx}{\linewidth}{@{}l l X X l c@{}}
    \toprule
    Suite & Type & \vary & \varo & \varm & $\vars{k}$ \\
    \midrule
    PICO       & Decomposable & Span sets (P/I/C/O) & Add/delete spans     & macro-F1 exact & $[0,1]\uparrow$ \\
    MedCalc    & Decomposable & Scalar (clearance)    & Perturb formula inputs & MSE            & $\geq 0\downarrow$ \\
    QA tasks   & Atomic       & Multiple-choice label & Swap wrong option      & boolean        & $\{0,1\}$ \\
    \bottomrule
  \end{tabularx}
  \caption{Task-specific operators, metrics, and ground-truth types (Algorithms~1 and~2). MedQA and PubMedQA use the same atomic row as MedBullets.}
  \label{tab:synth-by-suite}
\end{table}

Once candidates are ranked, Algorithm~2 bins all valid chosen--rejected pairs for a source example.
$\delta$-bin boundaries depend on $K$ and the number of bins~$N$ (even splits of rank gaps in $\{1, \ldots, K{-}1\}$; we use $N{=}3$ tertiles); scores enter only via the gap threshold $\tau{=}0.05$.
On atomic tasks ($m{=}\mathrm{boolean}$), every pair contrasts $C[1]$ with one lower rank and is placed in a single bin; on decomposable tasks, each rank gap $\delta$ fixes its bin $b$ before any candidate index is visited, and scores only gate whether a pair is kept ($|S[i] - S[j]| > \tau$).
Each stored tuple is $(\id{chosen}, \id{rejected})$, where each side is a triple $(i, C[i], S[i])$ recording rank index, candidate, and score.
Downstream preprocessing then draws one training example per source from these bins.
More formally:

\hypertarget{alg:synth-pair}{}%
\label{alg:synth-pair}
\begin{algorithmbox}[Algorithm 2]
\begin{codebox}
\Procname{$\proc{Bin-Preference-Pairs}(C, S, m, K, N)$}
\zi \Comment $N$: number of $\delta$-difficulty bins
\li let $B$ be an array of empty arrays \Comment bins of $(\id{chosen}, \id{rejected})$ triples
\li $\tau \gets 0.05$ \Comment score-gap threshold
\li \If{$m \isequal \text{boolean}$}
\li     \Then
      let $B[1]$ be an empty array
\li         \For $j \gets 2$ \To $K$
\li             \Do
        $\id{chosen} \gets (1, C[1], S[1])$; $\id{rejected} \gets (j, C[j], S[j])$ \Comment Eq.~\eqref{eq:pair}
\li                 append $(\id{chosen}, \id{rejected})$ to $B[1]$
  \End
\li \Else
      compute $\delta$-bin boundaries from $K$ and $N$
\li         \For $b \gets 1$ \To $N$
\li             \Do let $B[b]$ be an empty array
  \End
\li         \For $\delta \gets 1$ \To $K - 1$ \Comment Eq.~\eqref{eq:rank-gap}
\li             \Do
        $b \gets \delta$-bin of $\delta$ given boundaries
\li                 \For $i \gets 1$ \To $K - \delta$
\li                     \Do
          $j \gets i + \delta$
\li                         \If{$|S[i] - S[j]| > \tau$}
\li                             \Then
            $\id{chosen} \gets (i, C[i], S[i])$; $\id{rejected} \gets (j, C[j], S[j])$ \Comment Eq.~\eqref{eq:pair}
\li                                 append $(\id{chosen}, \id{rejected})$ to $B[b]$
      \End
    \End
  \End
  \End
\li \Return $B$
\end{codebox}
\end{algorithmbox}

Algorithm~2 organizes all valid preference pairs into difficulty-stratified bins; exact training examples are then selected by drawing one tuple per source and balancing underrepresented bins across the corpus (fixed random seed for reproducibility).
When training prompts are assembled, the chosen and rejected candidates are randomly assigned to slots $A$ and $B$ (50\% swap rate), again under a fixed seed, so gold verdicts from Eq.~\eqref{eq:pair} remain reproducible.

Below we illustrate the full pipeline with one worked example per family: PICO, MedCalc, and MedBullets as the representative atomic QA task.

\mutationsubsection{PICO}{https://huggingface.co/datasets/bigbio/ebm_pico}{app:mutation-pico}

\mutationexample{10070173}{test}

\mutationlabel{Source text \varx}
\begin{mutationplain}
Comparison of budesonide Turbuhaler with budesonide aqua in the treatment of seasonal allergic rhinitis. Rhinocort Study Group. OBJECTIVE To compare the effect of budesonide Turbuhaler 400 microg/day with budesonide aqua 256 microg/day in the treatment of seasonal allergic rhinitis (SAR). Secondarily to ascertain patients' preferences for the two nasal devices and to assess quality of life. DESIGN Randomized, multicentre, double-blind, double-dummy, parallel groups study. SETTING Private practices and hospital clinics in Ontario, Quebec and Manitoba. POPULATION Two hundred and eighty-four out-patients with SAR, who were symptomatic during the ragweed season, volunteered for enrolment (243 randomized). RESULTS Mean daily nasal symptom scores were significantly reduced with treatment. There were no statistically significant changes from baseline for eye symptoms. Most patients (more than 80\%) achieved substantial control of their symptoms with budesonide. The most common nasal and non-nasal adverse events for both groups were epistaxis and headache. Turbuhaler was easier to use and more convenient to carry, had less of an unpleasant taste, and caused less nasal irritation than the aqua spray. More than twice as many patients preferred Turbuhaler to the aqua spray (69\% versus 31\%). Improvement in quality of life from baseline to clinic visits was statistically significant in both groups. CONCLUSION Once daily use of 256 mg of budesonide aqua and 400 mg of budesonide Turbuhaler are equally safe and efficacious in the treatment of SAR. Patients preferred the budesonide powder formulation delivered via Turbuhaler two to one over the aqua formulation.
\end{mutationplain}

\mutationlabel{Ground truth \vary}
\begin{mutationplain}
\picoline{P}{\picoq{seasonal allergic rhinitis .}, \picoq{Two hundred and eighty-four}, \picoq{out-patients}, \picoq{SAR}, \picoq{symptomatic during the ragweed season}, \picoq{243}, \picoq{SAR .}}
\picoline{I}{\picoq{budesonide Turbuhaler}, \picoq{budesonide aqua}, \picoq{budesonide .}, \picoq{Turbuhaler}, \picoq{aqua spray}, \picoq{budesonide powder}, \picoq{aqua}}
\picoline{C}{(none)}
\picoline{O}{\picoq{patients ' preferences}, \picoq{quality of life}, \picoq{.}, \picoq{Mean daily nasal symptom scores}, \picoq{eye symptoms .}, \picoq{nasal and non-nasal adverse events}, \picoq{epistaxis}, \picoq{headache .}, \picoq{nasal irritation}, \picoq{safe}, \picoq{efficacious}}
\end{mutationplain}

\mutationlabel{Candidate $A$}
\begin{mutationplain}
\picoline{P}{\diffadd{\picoq{out-patients with SAR}}, \picoq{seasonal allergic rhinitis}, \diffadd{\picoq{ragweed season}}, \diffdel{\picoq{Two hundred and eighty-four}}, \diffdel{\picoq{out-patients}}, \diffdel{\picoq{SAR}}, \diffdel{\picoq{symptomatic during the ragweed season}}, \diffdel{\picoq{243}}, \diffdel{\picoq{SAR .}}}
\picoline{I}{\diffadd{\picoq{budesonide Turbuhaler 400 microg/day}}, \diffadd{\picoq{budesonide powder formulation}}, \picoq{Turbuhaler}, \diffdel{\picoq{budesonide Turbuhaler}}, \diffdel{\picoq{budesonide aqua}}, \diffdel{\picoq{budesonide .}}, \diffdel{\picoq{aqua spray}}, \diffdel{\picoq{budesonide powder}}, \diffdel{\picoq{aqua}}}
\picoline{C}{\diffadd{\picoq{budesonide aqua 256 microg/day}}, \diffadd{\picoq{aqua spray}}, \diffadd{\picoq{aqua formulation}}}
\picoline{O}{\diffadd{\picoq{nasal symptom scores}}, \picoq{eye symptoms}, \diffadd{\picoq{control of symptoms}}, \diffadd{\picoq{adverse events}}, \picoq{epistaxis}, \picoq{headache}, \diffadd{\picoq{ease of use}}, \diffadd{\picoq{convenient to carry}}, \diffadd{\picoq{unpleasant taste}}, \picoq{nasal irritation}, \picoq{patients' preferences}, \picoq{quality of life}, \picoq{safe}, \picoq{efficacious}, \diffdel{\picoq{.}}, \diffdel{\picoq{Mean daily nasal symptom scores}}, \diffdel{\picoq{nasal and non-nasal adverse events}}}
\end{mutationplain}

\mutationlabel{Candidate $B$}
\begin{mutationplain}
\picoline{P}{\diffadd{\picoq{out-patients with SAR}}, \picoq{seasonal allergic rhinitis}, \diffadd{\picoq{ragweed season}}, \diffdel{\picoq{Two hundred and eighty-four}}, \diffdel{\picoq{out-patients}}, \diffdel{\picoq{SAR}}, \diffdel{\picoq{symptomatic during the ragweed season}}, \diffdel{\picoq{243}}, \diffdel{\picoq{SAR .}}}
\picoline{I}{\diffadd{\picoq{budesonide Turbuhaler 400 microg/day}}, \diffadd{\picoq{budesonide powder formulation}}, \diffadd{\picoq{Turbuhaler device}}, \diffdel{\picoq{budesonide Turbuhaler}}, \diffdel{\picoq{budesonide aqua}}, \diffdel{\picoq{budesonide .}}, \diffdel{\picoq{Turbuhaler}}, \diffdel{\picoq{aqua spray}}, \diffdel{\picoq{budesonide powder}}, \diffdel{\picoq{aqua}}}
\picoline{C}{\diffadd{\picoq{budesonide aqua 256 microg/day}}, \diffadd{\picoq{aqua spray}}, \diffadd{\picoq{aqua formulation}}}
\picoline{O}{\diffadd{\picoq{nasal symptom scores}}, \picoq{eye symptoms}, \diffadd{\picoq{control of symptoms}}, \diffadd{\picoq{adverse events}}, \picoq{epistaxis}, \picoq{headache}, \diffadd{\picoq{ease of use}}, \diffadd{\picoq{convenience}}, \diffadd{\picoq{taste}}, \picoq{nasal irritation}, \picoq{patients' preferences}, \picoq{quality of life}, \picoq{safe}, \picoq{efficacious}, \diffdel{\picoq{.}}, \diffdel{\picoq{Mean daily nasal symptom scores}}, \diffdel{\picoq{nasal and non-nasal adverse events}}}
\end{mutationplain}

\mutationlabel{Scoring with macro-F1 exact match \varm\ against \vary}
\noindent Macro-F1 exact match lowercases spans and strips punctuation before testing equality, then scores set overlap per PICO key.
For each key $k \in \{P,I,C,O\}$, treat $\mathcal{Y}_k$ and the candidate spans $\mathcal{C}_k$ as sets; count TP, FP, and FN, then $P_k = \mathrm{TP}/(\mathrm{TP}+\mathrm{FP})$, $R_k = \mathrm{TP}/(\mathrm{TP}+\mathrm{FN})$, and $F1_k = 2P_k R_k/(P_k+R_k)$.

\noindent Candidate $A$:
\begin{itemize}
  \item $P$: $\mathrm{TP}{=}1$, $\mathrm{FP}{=}2$, $\mathrm{FN}{=}6$ $\Rightarrow$ $P{=}0.33$, $R{=}0.14$, $F1{=}0.20$
  \item $I$: $\mathrm{TP}{=}1$, $\mathrm{FP}{=}2$, $\mathrm{FN}{=}6$ $\Rightarrow$ $P{=}0.33$, $R{=}0.14$, $F1{=}0.20$
  \item $C$: $\mathrm{TP}{=}0$, $\mathrm{FP}{=}3$, $\mathrm{FN}{=}0$ $\Rightarrow$ $P{=}0.00$, $R{=}0.00$, $F1{=}0.00$
  \item $O$: $\mathrm{TP}{=}8$, $\mathrm{FP}{=}6$, $\mathrm{FN}{=}3$ $\Rightarrow$ $P{=}0.57$, $R{=}0.73$, $F1{=}0.64$
\end{itemize}
\noindent$\varm(A, \vary) = \frac{1}{4}\sum_k F1_k = 0.26$.

\noindent Candidate $B$:
\begin{itemize}
  \item $P$: $\mathrm{TP}{=}1$, $\mathrm{FP}{=}2$, $\mathrm{FN}{=}6$ $\Rightarrow$ $P{=}0.33$, $R{=}0.14$, $F1{=}0.20$
  \item $I$: $\mathrm{TP}{=}0$, $\mathrm{FP}{=}3$, $\mathrm{FN}{=}7$ $\Rightarrow$ $P{=}0.00$, $R{=}0.00$, $F1{=}0.00$
  \item $C$: $\mathrm{TP}{=}0$, $\mathrm{FP}{=}3$, $\mathrm{FN}{=}0$ $\Rightarrow$ $P{=}0.00$, $R{=}0.00$, $F1{=}0.00$
  \item $O$: $\mathrm{TP}{=}8$, $\mathrm{FP}{=}6$, $\mathrm{FN}{=}3$ $\Rightarrow$ $P{=}0.57$, $R{=}0.73$, $F1{=}0.64$
\end{itemize}
\noindent$\varm(B, \vary) = 0.21$.

\mutationlabel{Pair}
\noindent Chosen: $A$ ($s=0.26$);\quad Rejected: $B$ ($s=0.21$).

\mutationsubsection{MedCalc}{https://huggingface.co/datasets/ncbi/MedCalc-Bench-v1.2}{app:mutation-medcalc}

\mutationexample{8}{test}

\mutationlabel{Source text \varx}
\begin{mutationplain}
A 71-year-old man, non-smoker (height: 155 cm; weight: 59 kg), was scheduled to receive video-assisted thoracoscopic extended thymectomy with the diagnosis of MG. Two months previously, he developed symptoms of right ptosis and progressive swallowing difficulty. Based on a positive response to edrophonium and increased titers of autoantibodies to acetylcholine receptor (19.3 nmol/L; normal < 0.2 nmol/L), he was diagnosed as having MG with severity belonging to Osserman's classification IIb (ie, generalized moderate weakness and/or bulbar dysfunction). Thoracic computed tomography demonstrated glandular hyperplasia of the thymus (Fig. A). The patient was started on prednisolone 20 mg daily and pyridostigmine 60 mg three times daily. His past history included hypertension without evidence of previous myasthenic crisis or thromboembolic events (eg, history of lower limb swelling). The results of electrocardiography, pulmonary function test [eg, vital capacity: 93\%], echocardiography (eg, left ventricular ejection fraction: 85.1\%), chest radiography (Fig. B), and laboratory studies (eg, coagulation test) were unremarkable. On the other hand, impaired renal function [i.e., serum creatinine: 1.42 mg/dL; eGFR: 49.1 mL/min/1.73 m2] was observed after admission. Preoperative physical examination of the patient showed clear consciousness without respiratory distress. Vital signs included a blood pressure of 187/103 mm Hg, heart rate of 82 beats/min, and respiratory rate of 14 breaths/minute. Under real-time neuromuscular monitoring with a train-of-four (TOF) monitor (TOF-watch SX, N. V. Organon, Oss, Netherlands), anesthesia was induced with propofol (130 mg) and rocuronium (0.85 mg/kg). Following successful tracheal intubation with a double-lumen tracheal tube (Broncho-Cath; Mallinckrodt, Athlone, Ireland), general anesthesia was maintained with sevoflurane, rocuronium (total dosage: 40 mg), and a continuous infusion of remifentanil. An 18-gauge peripheral intravenous line and an arterial line were introduced. The surgical time was 4 hours 15 minutes with an estimated blood loss of 100 mL. Upon completion of surgery, sugammadex 4 mg/kg was administered to reverse neuromuscular blockade, with a maximum TOF ratio of 0.93 following reversal. Additionally, intravenous morphine 8 mg was given for postoperative analgesia. After successful extubation in the operating room and resumption of spontaneous breathing, he was transferred to the post-anesthesia care unit (PACU) for further care. During the immediate postoperative period, he was hemodynamically stable without respiratory distress. Because of surgical pain with a numeric rating scale of 5 (scale of 0--10), intravenous morphine was titrated to a total dosage of 7 mg. Forty-five minutes later, respiratory distress with drowsiness was noted. Physical examination found pinpoint pupils with a TOF ratio of 0.9. Blood gas analysis demonstrated severe hypercapnia (arterial carbon dioxide pressure: 117.7 mm Hg) and acidosis (pH: 6.996, lactate levels: 3.3 mmol/L). On suspicion of morphine overdose, intravenous naloxone was administered twice (0.08 mg each time). After 20 minutes, the patient regained consciousness and normal respiratory pattern. Subsequent blood gas analysis demonstrated hyperlactatemia (lactate levels: 6.0 mmol/L) (Fig.) despite improvement in arterial carbon dioxide pressure and arterial oxygen pressure after naloxone administration. Taking into account the overall clinical improvement, he was transferred to ward with spontaneous breathing and stable hemodynamics after observation for 100 minutes following initial reversal of opioid overdose. However, one hour after being transferred to ward (ie, three hours after naloxone administration), he was found to exhibit consciousness loss, respiratory distress, and pinpoint pupils. Intravenous naloxone 0.4 mg was given, followed by endotracheal intubation and transfer to the intensive care unit for mechanical ventilatory support. Brain computed tomography showed no intracranial lesion. He regained consciousness two hours after naloxone administration in the intensive care unit without symptoms of opioid withdrawal (eg, pulmonary edema). As weaning from mechanical ventilation was difficult on POD 3, a diagnosis of POMC was made. Steroid therapy (prednisolone 40 mg twice daily) and pyridostigmine (60 mg 3 times a day) was initiated. The patient was extubated smoothly on POD 6 and was discharged from hospital on POD 12. The course of hyperlactatemia was shown in Figure. In addition, his eGFR increased from 49.1 mL/min/1.73m2 at baseline to 88.9 ml/min/1.73m2 on POD 5. Pathological analysis of the specimen from thymectomy confirmed the diagnosis of type B2 thymoma according to the World Health Organization (WHO) classification. The patient was readmitted on POD 24 because of left thigh swelling. Ultrasonographic examination showed evidence of DVT involving the left femoral and popliteal veins. Anticoagulant therapy with low-molecular-weight heparin (ie, subcutaneous Clexane 60 mg every 12 hours) was implemented immediately aflter hospitalization, and he was discharged without sequelae on POD 31. During hospitalization, his eGFR was 86.5 mL/min/1.73m2. There was no recurrence of myasthenic crisis or DVT up to 3 months of follow-ups.What is the patient's Creatinine Clearance using the Cockroft-Gault Equation in terms of mL/min? You should use the patient's adjusted body weight in kg instead of the patient's actual body weight if the patient is overweight or obese based on their BMI. If the patient's BMI's normal, set their adjusted body weight to the minimum of the ideal body and actual weight. If the patient is underweight, please set their adjusted body weight to their actual body weight. You should use the patient's medical values and health status when they were first admitted to the hospital prior to any treatment.
\end{mutationplain}

\mutationlabel{Ground truth \vary}
\begin{mutationplain}
\argline{age}{71}
\argline{sex}{male}
\argline{adjusted\_body\_weight\_kg}{52.35}
\argline{serum\_creatinine\_mg\_dL}{1.42}
\argline{clearance\_mL\_min}{35.33}
\end{mutationplain}

\mutationlabel{Candidate $A$}
\begin{mutationplain}
\argline{age}{71}
\argline{sex}{male}
\argline{adjusted\_body\_weight\_kg}{\diffdel{52.35}\diffadd{52.7}}
\argline{serum\_creatinine\_mg\_dL}{1.42}
\end{mutationplain}

\mutationlabel{Candidate $B$}
\begin{mutationplain}
\argline{age}{71}
\argline{sex}{male}
\argline{adjusted\_body\_weight\_kg}{52.35}
\argline{serum\_creatinine\_mg\_dL}{1.42}
\end{mutationplain}

\mutationlabel{Scoring with MSE \varm\ against \vary}
\noindent Cockcroft--Gault:
\par\vspace{2pt}
\noindent{\small\ttfamily\frenchspacing
CrCl(age, sex, adjusted\_body\_weight\_kg, serum\_creatinine\_mg\_dL) =\\
\quad ((140 - age) * adjusted\_body\_weight\_kg * (1 if sex == 'male' else 0.85))\\
\quad / (serum\_creatinine\_mg\_dL * 72)\par\vspace{4pt}}
\noindent$\varm(c, \vary) = \mathrm{mean}\bigl((\mathrm{CrCl}(c) - \vary)^2\bigr)$, with a single scalar clearance target.

\noindent Candidate $A$:
\par\vspace{2pt}
\noindent{\small\ttfamily\frenchspacing
CrCl(71, "male", 52.7, 1.42)\\
\quad = ((140 - 71) * 52.7 * 1) / (1.42 * 72)\\
\quad = 35.57\par}
\noindent$\varm(A, \vary) = \mathrm{mean}\bigl((35.57 - 35.33)^2\bigr) = \mathrm{mean}(0.23^2) = 0.054$.

\noindent Candidate $B$:
\par\vspace{2pt}
\noindent{\small\ttfamily\frenchspacing
CrCl(71, "male", 52.35, 1.42)\\
\quad = ((140 - 71) * 52.35 * 1) / (1.42 * 72)\\
\quad = 35.33\par}
\noindent$\varm(B, \vary) = \mathrm{mean}\bigl((35.33 - 35.33)^2\bigr) = 0.0$.

\mutationlabel{Pair}
\noindent Chosen: $B$ ($s=0.0$);\quad Rejected: $A$ ($s=0.054$).

\mutationsubsection{MedBullets}{https://huggingface.co/datasets/super-dainiu/medagents-benchmark}{app:mutation-medbullets}

\mutationexample{0}{MedBullets/train}

\mutationlabel{Source text \varx}
\begin{mutationplain}
A 64-year-old man presents to the emergency room with a headache and nausea. He reports that he was rocking his grandson to sleep when the symptoms began. He states the pain is constant and is primarily located on his right side. When asked to indicate the area of pain, he says that it surrounds his eye and upper forehead. He had one episode of vomiting. The patient also reports difficulty seeing out of his right eye, which he attributes to excessive tearing. The patient's past medical history is significant for hypertension. His medications include hydrochlorothiazide. His temperature is 98.6°F (37°C), blood pressure is 135/91 mmHg, pulse is 72/min, and respirations are 12/min. The patient's right eye is shown in Figure A. Upon physical examination, the right pupil is minimally responsive to light and the globe feels firm. A right-sided carotid bruit is appreciated. Which of the following is the most appropriate prophylaxis for this patient's condition? Options: A. Acetazolamide B. Amitriptyline C. Clopidogrel D. Epinephrine E. Verapamil
\end{mutationplain}

\mutationlabel{Ground truth \vary}
\begin{mutationplain}
A. Acetazolamide
\end{mutationplain}

\mutationlabel{Candidate $A$}
\begin{mutationplain}
A. Acetazolamide
\end{mutationplain}

\mutationlabel{Candidate $B$}
\begin{mutationplain}
\diffdel{A. Acetazolamide}\diffadd{C. Clopidogrel}
\end{mutationplain}

\mutationlabel{Scoring with boolean correctness \varm\ against \vary}
\noindent$\varm(c, \vary) = 1.0$ if $c$ exactly matches $\vary$, else $0.0$.

\noindent Candidate $A$:
\noindent$\varm(A, \vary) = 1.0$ if $c{=}\vary$ else $0.0 = 1.0$.

\noindent Candidate $B$:
\noindent$\varm(B, \vary) = 1.0$ if $c{=}\vary$ else $0.0 = 0.0$.

\mutationlabel{Pair}
\noindent Chosen: $A$ ($s=1.0$);\quad Rejected: $B$ ($s=0.0$).

\section{Prompt templates}
\label{app:prompts}

The prompts below are adapted from the J1~\cite{j1} thinking seed templates. We reuse them throughout the pipeline: when generating distillation data for SFT, when tuning with SFT or RL, and finally when meta-evaluating.
Every judge prompt has a system message:
\label{app:system}
\begin{promptblock}
\begin{promptverbatim}
You are an impartial judge evaluating how well response(s) fulfill the user's instructions.
\end{promptverbatim}
\end{promptblock}
and a user message composed conceptually of four blocks: 1) \emph{format instructions}, 2) \emph{task instructions} (also called ``rubric''), 3) \emph{source text}, and 4) \emph{candidate response(s)}.
The exact contents of the user message template vary by judge mode, since we need different format instructions for pairwise or pointwise, and by task, as task instructions are idiosyncratic. We detail each below.

\subsection{Pairwise}
\label{app:pairwise}

\begin{promptblock}
\begin{promptverbatim}
You are given a user question and two responses from two AI assistants. Your task is to act as an impartial judge and evaluate which response better follows the user's instructions and provides a higher-quality answer.

Think carefully about how to assess the quality of the responses, and enclose your reasoning within <think> and </think> tags. Your reasoning should include your evaluation criteria, a clear understanding of what an ideal response would look like for this particular question, and a concrete example of such an ideal or reference answer if possible. Then compare both assistants' responses to your ideal or reference answer, explaining how each aligns with or deviates from your expectations. Be specific and avoid vague or overly general judgments. Remain as objective as possible.

Finally, provide your verdict within <answer> and </answer> tags, strictly following this format:
- <answer> [[A]] </answer> if Assistant A is better
- <answer> [[B]] </answer> if Assistant B is better

Format your output like this:
<think> your_thinking_process </think>
<answer> [[A]] </answer>  (or [[B]])

Below are the user's question and the two responses:

[User Question]
{instruction}
{text}

[The Start of Assistant A's Answer]
{response_A}
[The End of Assistant A's Answer]

[The Start of Assistant B's Answer]
{response_B}
[The End of Assistant B's Answer]
\end{promptverbatim}
\end{promptblock}
\promptwhere
\begin{itemize}
  \item \texttt{\{instruction\}} --- task instructions (or ``rubric'') from \S\ref{app:rubric} (PICO, MedCalc, or QA).
  \item \texttt{\{text\}} --- source passage for the example (e.g., abstract, clinical note, or exam stem), from the task benchmark via \S\ref{sec:synthetic-data}.
  \item \texttt{\{response\_A\}} --- candidate output in slot~$A$ of a preference pair, from the mutation operator in \S\ref{sec:synthetic-data} (worked examples in Appendix~\ref{app:mutations}); slot assignment varies across training batches to mitigate position bias.
  \item \texttt{\{response\_B\}} --- candidate output in slot~$B$, same provenance as \texttt{\{response\_A\}}.
\end{itemize}

\subsection{Pointwise}
\label{app:pointwise}

\begin{promptblock}
\begin{promptverbatim}
You are given a user question and a response from an AI assistant. Your task is to act as an impartial judge and evaluate how well the response fulfills the user's instructions.

Think carefully about how to assess the quality of the response, and enclose your reasoning within <think> and </think> tags. Your reasoning should include your evaluation criteria, a clear understanding of what an ideal response would look like for this particular question, and a concrete example of such an ideal or reference answer if possible. Then compare the assistant's response to your ideal or reference answer, explaining how it aligns with or deviates from your expectations. Be specific and avoid vague or overly general judgments. Remain as objective as possible.

Finally, assign the assistant's response a score from 0 to 10, using either an integer or a decimal with up to 0.1 precision. A higher score should indicate a higher-quality response. Enclose the score within <score> and </score> tags.

Format your output like this:
<think> your_thinking_process </think>
<score> your_score </score>

Below are the user's question and the assistant's response:

[User Question]
{instruction}
{text}

[The Start of the Assistant's Answer]
{response}
[The End of the Assistant's Answer]
\end{promptverbatim}
\end{promptblock}
\promptwhere
\begin{itemize}
  \item \texttt{\{instruction\}} --- task instructions (or ``rubric'') from \S\ref{app:rubric} (PICO, MedCalc, or QA).
  \item \texttt{\{text\}} --- source passage for the example, same provenance as in the pairwise template.
  \item \texttt{\{response\}} --- a single candidate output from \S\ref{sec:synthetic-data}, presented one at a time for pointwise judging (or in distillation rows).
\end{itemize}

\anatomysubsection{anatRubric}{Task instructions}{app:rubric}

\promptvariant{PICO}
\label{app:task-pico}
\begin{promptblock}
\begin{promptverbatim}
Extract PICO elements from this biomedical text. Identify Population (P), Intervention (I), Comparison (C), and Outcome (O) components mentioned in the study. Output should be a JSON object with keys "P", "I", "C", and "O", where values are arrays of strings. Each string should be a verbatim copy of the relevant text from the original document. Keep each string as short as possible, containing only the topic words. Avoid including all connectors and other words that are not part of the topic. Numbers should be separated from the following words they refer to.
\end{promptverbatim}
\end{promptblock}

\promptvariant{MedCalc}
\label{app:task-medcalc}

\begin{promptblock}
\begin{promptverbatim}
You are given the generic formula, together with a dictionary containing the definitions of all variables. Find in the patient note the current values for these variables and substitute them in the formula, do measurements conversion if needed. All variables have to be specified. Also manipulate the formula to make it Python executable in a single line, with no imports or intermediate variables definitions. Do not try to calculate the results of the formula, just write the correct values of the variables. If working with dates or time deltas, use the Python library datetime which has been already imported as from datetime import datetime, timedelta. For mathematical functions, such as log, use the Python library numpy which has been already imported as import numpy as np. If a specific formatting is required, unless it is a simple number, specify it as if it had to be executed as a Python string. Make sure that all the specific variables needed to compute accurately the formula have been provided, otherwise signal it in the all_inputs flag. If the value of a categorical booelan variable can not be determined from the description, for instance something like Diabetes_history, assume it to be False and do not flag it as absent in the all_inputs flag. If the value of another kind of variable can not be determined, check how it is used in the formula. If it is used only in a comparison, such as age < 65, assume it to be a value that makes the comparison false. The selected formula {calculator} has the following definition: {formula}. The variables are defined as: {formula_inputs}. The output format has to be {out_format}.
\end{promptverbatim}
\end{promptblock}
\promptwhere
\begin{itemize}
  \item \texttt{\{calculator\}} --- name of the selected MedCalc-Bench formula for this example (from the benchmark metadata).
  \item \texttt{\{formula\}} --- mathematical definition of that formula (from the benchmark's formula catalog).
  \item \texttt{\{formula\_inputs\}} --- variable definitions and units required by the formula.
  \item \texttt{\{out\_format\}} --- required output format for the substituted, Python-evaluable formula string.
\end{itemize}

\promptvariant{QA (MedQA, MedBullets, PubMedQA)}
\label{app:task-medqa}
\begin{promptblock}
\begin{promptverbatim}
You are given a medical question and a set of possible answers. Select the most appropriate answer from the given options.
\end{promptverbatim}
\end{promptblock}

\section{Detailed results}
\label{app:tables}

Figure~\ref{fig:judging-outcomes} summarizes pairwise judging correctness, robustness, and compliance; Tables~\ref{tab:correctness},~\ref{tab:variance}, and~\ref{tab:invalid} report the exact per-suite values.
Figure~\ref{fig:rank-gap-difficulty} summarizes rank-gap difficulty on decomposable items; Table~\ref{tab:rank-gap} reports the exact per-tertile values.

\begin{table}[H]
  \centering
  \footnotesize
  \setlength{\tabcolsep}{3.5pt}
  \begin{tabular}{@{}l*{4}{c}@{\hspace{0.5em}}|@{\hspace{0.5em}}*{3}{c}@{}}
    \evaltabheader
    \textbf{Decomposable} ($n{=}1026$) & 0.75 & 0.79* \gdelta{+0.04} & 0.67* \rdelta{$-$0.08} & \textbf{0.84*} \gdelta{+0.09} & 0.82 & \underline{\textbf{0.87}} & 0.86 \\
    \quad PICO ($n{=}187$)      & 0.84 & 0.88 \gdelta{+0.04} & 0.78 \rdelta{$-$0.06} & \underline{\textbf{0.94*}} \gdelta{+0.10} & 0.86 & \underline{\textbf{0.94}} & \underline{\textbf{0.94}} \\
    \quad MedCalc ($n{=}839$)   & 0.65 & 0.71* \gdelta{+0.06} & 0.57* \rdelta{$-$0.08} & \textbf{0.75*} \gdelta{+0.10} & 0.79 & \underline{\textbf{0.80}} & 0.79 \\
    \textbf{Atomic} ($n{=}2081$)       & 0.76 & \textbf{0.79*} \gdelta{+0.03} & 0.69* \rdelta{$-$0.07} & 0.78 \gdelta{+0.02} & \underline{\textbf{0.91}} & \underline{\textbf{0.91}} & \underline{\textbf{0.91}} \\
    \quad MedBullets ($n{=}308$) & 0.69 & 0.72 \gdelta{+0.03} & 0.63* \rdelta{$-$0.06} & \textbf{0.73} \gdelta{+0.04} & \underline{\textbf{0.94}} & 0.93 & \underline{\textbf{0.94}} \\
    \quad MedQA ($n{=}1273$)    & 0.79 & \textbf{0.82*} \gdelta{+0.03} & 0.71* \rdelta{$-$0.08} & 0.78 \rdelta{$-$0.01} & 0.96 & \underline{\textbf{0.97}} & 0.96 \\
    \quad PubMedQA ($n{=}500$)  & 0.79 & 0.83* \gdelta{+0.04} & 0.74* \rdelta{$-$0.05} & \underline{\textbf{0.84*}} \gdelta{+0.05} & 0.83 & \underline{\textbf{0.84}} & \underline{\textbf{0.84}} \\
    \midrule
    \textbf{Avg}          & 0.75 & 0.79* \gdelta{+0.04} & 0.69* \rdelta{$-$0.06} & \textbf{0.81*} \gdelta{+0.06} & 0.87 & \underline{\textbf{0.89}} & \underline{\textbf{0.89}} \\
    \bottomrule
  \end{tabular}
  \caption{Pairwise judging \textbf{correctness} with rows labeled \textbf{Decomposable} and \textbf{Atomic} as unweighted family means, indented rows as individual benchmarks, \textbf{Avg} as the unweighted mean across all five suites, \textbf{Bold} as best-in-class among Llama-8B or frontier columns, underlined bold as overall row best, and asterisk as McNemar significance~\cite{mcnemar} vs.\ baseline ($p < 0.05$).}
  \label{tab:correctness}
\end{table}

\begin{table}[H]
  \centering
  \footnotesize
  \setlength{\tabcolsep}{3.5pt}
  \begin{tabular}{@{}l*{4}{c}@{\hspace{0.5em}}|@{\hspace{0.5em}}*{3}{c}@{}}
    \evaltabheader
    PICO ($n{=}187$)      & 0.14 & 0.06 \gdelta{$-$0.08} & \textbf{0.03} \gdelta{$-$0.11} & 0.06 \gdelta{$-$0.08} & 0.09 & 0.03 & \underline{\textbf{0.02}} \\
    MedCalc ($n{=}839$)   & 0.07 & 0.05 \gdelta{$-$0.02} & 0.03 \gdelta{$-$0.04} & \textbf{0.02} \gdelta{$-$0.05} & 0.10 & \underline{\textbf{0.01}} & 0.02 \\
    MedBullets ($n{=}308$) & 0.05 & \textbf{0.04} \gdelta{$-$0.01} & 0.06 \rdelta{+0.01} & 0.07 \rdelta{+0.02} & 0.11 & \underline{\textbf{0.01}} & \underline{\textbf{0.01}} \\
    MedQA ($n{=}1273$)    & \textbf{0.04} & 0.09 \rdelta{+0.05} & 0.09 \rdelta{+0.05} & \textbf{0.04} \gdelta{+0.00} & 0.07 & 0.01 & \underline{\textbf{0.00}} \\
    PubMedQA ($n{=}500$)  & 0.08 & \textbf{0.06} \gdelta{$-$0.03} & 0.11 \rdelta{+0.03} & \textbf{0.06} \gdelta{$-$0.02} & 0.03 & 0.01 & \underline{\textbf{0.00}} \\
    \midrule
    \textbf{Avg}          & 0.08 & 0.06 \gdelta{$-$0.02} & 0.07 \gdelta{$-$0.01} & \textbf{0.05} \gdelta{$-$0.03} & 0.08 & \underline{\textbf{0.01}} & \underline{\textbf{0.01}} \\
    \bottomrule
  \end{tabular}
  \caption{\textbf{Verdict variance} is the mean $|\mathrm{high}-\mathrm{low}|$ across datasets over majority-vote and any-/all-sample correctness rates (lower is better), with the same columns and formatting as Table~\ref{tab:correctness}.}
  \label{tab:variance}
\end{table}

\begin{table}[H]
  \centering
  \footnotesize
  \setlength{\tabcolsep}{3.5pt}
  \begin{tabular}{@{}l*{4}{c}@{\hspace{0.5em}}|@{\hspace{0.5em}}*{3}{c}@{}}
    \evaltabheader
    PICO ($n{=}187$)      & \underline{\textbf{0.0\%}} & 4.5\% \rdelta{+4.5} & \underline{\textbf{0.0\%}} \gdelta{+0.0} & \underline{\textbf{0.0\%}} \gdelta{+0.0} & 3.7\% & \underline{\textbf{0.0\%}} & \underline{\textbf{0.0\%}} \\
    MedCalc ($n{=}839$)   & 5.5\% & 6.5\% \rdelta{+1.0} & 4.7\% \gdelta{$-$0.8} & \textbf{4.2\%} \gdelta{$-$1.3} & \underline{\textbf{0.0\%}} & \underline{\textbf{0.0\%}} & \underline{\textbf{0.0\%}} \\
    MedBullets ($n{=}308$) & 1.1\% & \underline{\textbf{0.0\%}} \gdelta{$-$1.1} & 4.3\% \rdelta{+3.2} & \underline{\textbf{0.0\%}} \gdelta{$-$1.1} & \underline{\textbf{0.0\%}} & \underline{\textbf{0.0\%}} & \underline{\textbf{0.0\%}} \\
    MedQA ($n{=}1273$)    & \textbf{0.4\%} & 0.9\% \rdelta{+0.5} & 3.8\% \rdelta{+3.4} & 0.7\% \rdelta{+0.3} & 12.8\% & \underline{\textbf{0.0\%}} & \underline{\textbf{0.0\%}} \\
    PubMedQA ($n{=}500$)  & \underline{\textbf{0.0\%}} & \underline{\textbf{0.0\%}} \gdelta{+0.0} & \underline{\textbf{0.0\%}} \gdelta{+0.0} & \underline{\textbf{0.0\%}} \gdelta{+0.0} & 1.2\% & \underline{\textbf{0.0\%}} & \underline{\textbf{0.0\%}} \\
    \midrule
    \textbf{Avg}          & 2.3\% & 2.8\% \rdelta{+0.5} & 3.6\% \rdelta{+1.3} & \textbf{1.6\%} \gdelta{$-$0.7} & 2.2\% & \underline{\textbf{0.0\%}} & \underline{\textbf{0.0\%}} \\
    \bottomrule
  \end{tabular}
  \caption{\textbf{Invalid-response share} is invalid / (invalid + legitimate errors), lower is better, with the same columns and formatting as Table~\ref{tab:correctness}.}
  \label{tab:invalid}
\end{table}

\begin{table}[H]
  \centering
  \footnotesize
  \setlength{\tabcolsep}{3.5pt}
  \begin{tabular}{@{}l*{4}{c}@{\hspace{0.5em}}|@{\hspace{0.5em}}*{3}{c}@{}}
    \evaltabheader
    \textbf{Hard} ($n{=}439$)   & 0.62 & 0.65 \gdelta{+0.03} & 0.54* \rdelta{$-$0.08} & \textbf{0.69*} \gdelta{+0.07} & 0.69 & \underline{\textbf{0.72}} & \underline{\textbf{0.72}} \\
    \textbf{Medium} ($n{=}440$) & 0.72 & 0.79* \gdelta{+0.07} & 0.63* \rdelta{$-$0.09} & \textbf{0.83*} \gdelta{+0.11} & 0.86 & \underline{\textbf{0.88}} & 0.87 \\
    \textbf{Easy} ($n{=}147$)   & 0.79 & 0.86 \gdelta{+0.07} & 0.72 \rdelta{$-$0.07} & \textbf{0.90*} \gdelta{+0.11} & 0.94 & 0.94 & \underline{\textbf{0.95}} \\
    \bottomrule
  \end{tabular}
  \caption{\textbf{Rank-gap difficulty} on decomposable items (PICO + MedCalc), where \textbf{Hard} / \textbf{Medium} / \textbf{Easy} are three tertiles of $\delta$ per source example with boundaries over $[1, K{-}1]$ that depend on that example's candidate count $K$ (hard = smallest gaps), not fixed absolute integers across tasks, using the same columns and formatting as Table~\ref{tab:correctness}.}
  \label{tab:rank-gap}
\end{table}

\end{document}